\pdfoutput=1

\documentclass[11pt]{article}

\usepackage[]{acl}

\usepackage{times}
\usepackage{latexsym}

\usepackage[T1]{fontenc}

\usepackage[utf8]{inputenc}

\usepackage{microtype}

\usepackage{inconsolata}

\usepackage{graphicx}
\usepackage{amssymb}
\usepackage{amsmath}
\usepackage{enumitem}
\usepackage{booktabs}
\usepackage{multirow}
\usepackage{bbm}
\usepackage{dsfont}

\title{Efficient Prompting for Continual Adaptation to Missing Modalities}

\author{Zirun Guo, Shulei Wang, Wang Lin, Weicai Yan, Yangyang Wu\thanks{\quad Corresponding author} , Tao Jin\\
Zhejiang University\\
\texttt{zrguo.cs@gmail.com}
}

\begin{document}
\maketitle
\begin{abstract}
  Missing modality issues are common in real-world applications, arising from factors such as equipment failures and privacy concerns. When fine-tuning pre-trained models on downstream datasets with missing modalities, performance can degrade significantly. Current methods often aggregate various missing cases to train recovery modules or align multimodal features, resulting in suboptimal performance, high computational costs, and the risk of catastrophic forgetting in continual environments where data arrives sequentially.
  In this paper, we formulate the dynamic missing modality problem as a continual learning task and introduce the continual multimodal missing modality task. To address this challenge efficiently, we introduce three types of prompts: modality-specific, task-aware, and task-specific prompts. These prompts enable the model to learn intra-modality, inter-modality, intra-task, and inter-task features. Furthermore, we propose a contrastive task interaction strategy to explicitly learn prompts correlating different modalities.
  We conduct extensive experiments on three public datasets, where our method consistently outperforms state-of-the-art approaches.
\end{abstract}

\section{Introduction}
\begin{figure}
  \centering
  \includegraphics[width=\linewidth]{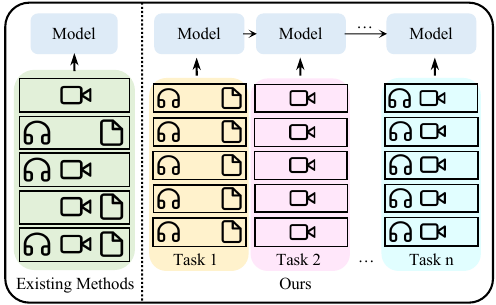}
  \caption{The difference between existing methods and ours. Existing methods train all cases of data together, which is infeasible in many real-world scenarios. In contrast, we formulate it as a continual learning problem, which is much closer to real situations.}
  \label{problem}
  \vskip -0.15 in
\end{figure}

Pre-trained multimodal models have shown great potential in many applications~\citep{radford2021learning, Li2023BLIP2BL, lin2024action}. When fine-tuning these pre-trained models on downstream tasks, missing modality issues often occur due to equipment failure, data corruption, privacy concerns, etc. Existing methods~\citep{ma2021smil, zhao-etal-2021-missing, lee2023multimodal, guo-etal-2024-multimodal} address missing modality issues by reconstructing missing information or aligning multimodal features. However, both recovering missing features and aligning multimodal features are based on datasets containing various types of missing modality cases (Figure~\ref{problem} (left)) to achieve robust performance. For example, recovering feature methods learn how to reconstruct a missing modality using the available modalities. Hence, it is expected that the dataset contains various types of missing cases to optimize the reconstruction modules.

\begin{figure}
  \centering
  \includegraphics[width=\linewidth]{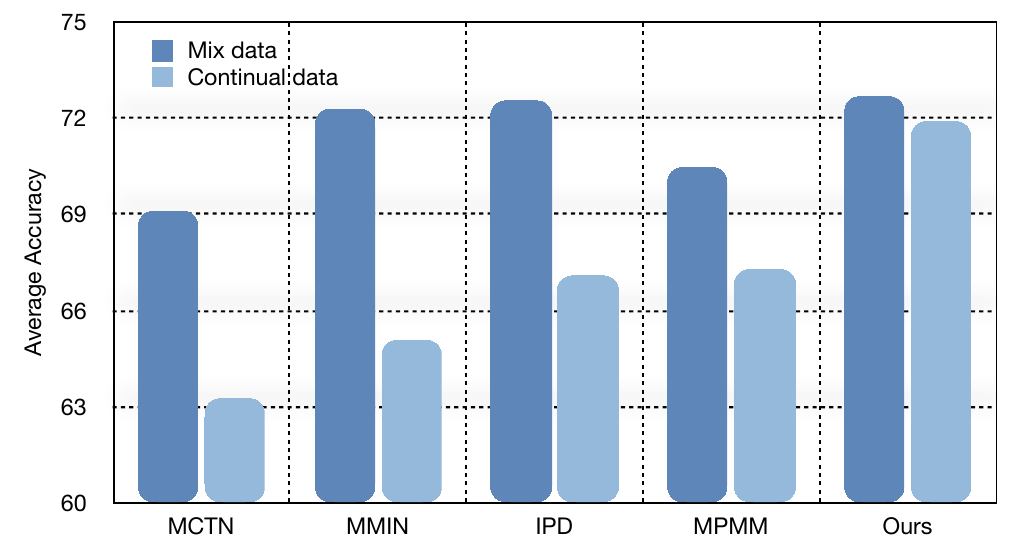}
  \caption{The performance of existing methods will degrade when applied to continual multimodal missing modality task.}
  \label{comp}
  \vskip -0.15 in
\end{figure}

However, in real-world dynamic environments, data often comes in a sequence where each dataset has the same modality missing (Figure~\ref{problem} (right)). For example, a robot needs to utilize multiple sensors to capture human faces, gestures and speech to analyze sentiment and emotion. When the camera is broken, the system needs to make predictions without video modality during the period until the camera is repaired. During this period, all the data has the same missing modality (\textit{i.e.} video). When the recording device is broken, the system needs to learn how to make accurate predictions without audio modality until the recording device is repaired. In such dynamic environments, the system is expected to adapt to the different missing modality cases continually. Therefore, existing methods relying on recovering missing features and aligning multimodal features will fail. Additionally, as shown in Figure~\ref{comp}, the sequential data in real-world applications will make these methods suffer catastrophic forgetting~\citep{mccloskey1989catastrophic}, leading to performance degradation. To handle forgetting, an intuitive idea is to store and retrain all old data but it creates large storage overheads and potential privacy issues.

Based on the above observations, we propose the continual multimodal missing modality task to address the missing modality issues in real-world continual environments. In recent years, continual learning has made great progress, such as replay-based methods~\citep{rolnick2019experience, buzzega2020dark, cha2021co2l}, regularization-based methods~\citep{kirkpatrick2017overcoming, zenke2017continual, aljundi2018memory}, and architecture-based methods~\citep{serra2018overcoming, li2019learn, ebrahimi2020adversarial}. However, these methods often have many limitations. For example, replay-based methods need to store previous data, which could pose potential privacy issues.
More recently, prompt-based continual methods~\citep{wang2022s, wang2022learning, wang2022dualprompt} have attracted much attention due to their simplicity and effectiveness. Most of these methods~\citep{wang2022learning, wang2022dualprompt} are unimodal and are difficult to transfer to the multimodal field. Multimodal methods~\citep{wang2022s, qian2023decouple} always depend on language-image models such as CLIP~\citep{radford2021learning}, which makes it difficult to apply these methods to other fields where there are more modalities. Moreover, these multimodal methods focus more on exploring task interaction while ignoring modality interaction.

In this paper, we propose three types of prompts and a task interaction strategy for efficient continual multimodal missing modality task. Specifically, we propose modality-specific prompts, task-aware prompts, and task-specific prompts. Modality-specific prompts aim to instruct the model to learn intra-modality features. Task-aware prompts focus on learning inter-modality and inter-task features. Task-specific prompts help the model learn intra-task features. Moreover, we propose a contrastive task interaction strategy to grasp the relationships between tasks.

We conduct extensive experiments on three multimodal datasets: CMU-MOSI~\citep{7742221}, IEMOCAP~\citep{Busso2008IEMOCAPIE} and CH-SIMS~\cite{yu-etal-2020-ch}. Our proposed method can consistently outperform baselines and state-of-the-art methods significantly in all three datasets. Besides, the number of trainable parameters only accounts for 2-3\% of the parameters of the backbone network, indicating our method is parameter-efficient. We further conduct ablation experiments to validate the effectiveness of three types of prompts and contrastive task interaction strategy. The results fully demonstrate the superiority of our method. Our main contributions can be summarized as follows:
\begin{itemize}
  \item We introduce a comprehensive formulation of \textit{continual multimodal missing modality} task.
  \item We propose modality-specific prompts, task-aware prompts, task-specific prompts and a contrastive task interaction strategy. They can be transferred easily to any multimodal backbones efficiently.
  \item We build up three benchmarks for continual multimodal missing modalities. Our proposed method outperforms all the baselines and state-of-the-art approaches significantly.
\end{itemize}

\begin{figure*}
  \centering
  \includegraphics[width=0.96\linewidth]{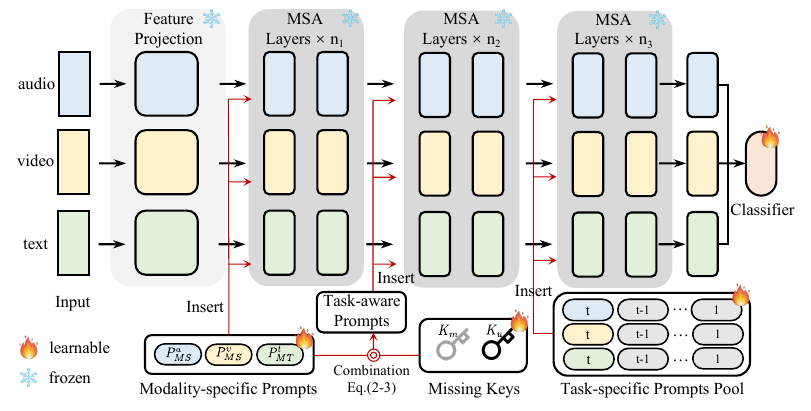}
  \caption{The overall architecture of our proposed method. After the projection layer, modality-specific prompts, task-aware prompts and task-specific prompts are attached to multi-head self-attention (MSA) layers sequentially. Task-aware prompts are generated from modality-specific prompts and missing keys using Eq.(\ref{e2}).}
  \label{overall}
  \vskip -0.1 in
\end{figure*}

\section{Related Work}
\noindent\textbf{Multimodal Learning with Missing Modalities.}
Missing modality issues pose challenges for multimodal learning~\citep{guo2024classifier} and can lead to severe performance degradation. Recently, many works explore to address the missing modality issues~\citep{ma2021smil, 10.1145/3219819.3219963, Du_2018, zhao-etal-2021-missing,lee2023multimodal,10.1145/3581783.3612291}. Some methods~\citep{10.1145/3219819.3219963, Du_2018} directly generate missing modalities using the available modalities. \citet{pham2020translation} propose to align multimodal features by translating between modalities to address missing modality issues. \citet{zhao-etal-2021-missing} propose learning robust joint multimodal representations that can predict the representation of any missing modality given the available modalities. IPD~\citep{10.1145/3581783.3612291} jointly learns modality-specific task prototypes. \citet{guo-etal-2024-multimodal} propose three types of prompts to address missing modality issues in a parameter-efficient way. \citet{guo2025smoothing} propose to address missing modalities at test time by smoothing the distribution shifts between the complete data samples and incomplete data samples.

\noindent\textbf{Continual Learning.}
A major challenge of continual learning is known as catastrophic forgetting~\citep{mccloskey1989catastrophic}. Numerous methods have been exploited to address this issue which could be categorized into three main categories: (1) Regularization-based approaches~\citep{kirkpatrick2017overcoming, zenke2017continual, aljundi2018memory} address catastrophic forgetting by imposing a regularization constraint to important parameters. 
(2) Replay-based approaches~\citep{rebuffi2017icarl, shin2017continual, rolnick2019experience, buzzega2020dark, cha2021co2l} store some representative samples of previous tasks in a rehearsal buffer and retrain these data to avoid forgetting. 
(3) Architecture-based approaches~\citep{mallya2018packnet, serra2018overcoming, li2019learn, ebrahimi2020adversarial} dynamically expand or divide the network for different tasks to mitigate forgetting. These methods often suffer from scalability issues as the number of tasks or the complexity of the model increases. Our proposed method is based on prompt learning and is a replay-free method. Moreover, our novel design of prompts can instruct the model to address complex situations compared to regularization-based methods and architecture-based methods.

\noindent\textbf{Prompt Learning.}
Prompt learning, as one of the efficient transfer learning techniques~\citep{hu2021lora, guo2024wander, yan2025diffprompt}, refers to the process of designing or generating effective prompts to use a pre-trained model for different types of downstream tasks. Recent works~\citep{wang2022s, wang2022learning, wang2022dualprompt, yan2024lowrank} apply prompt learning to the field of continual learning and have achieved good results. DualPrompt~\citep{wang2022dualprompt} proposes G-Prompt and E-Prompt to learn task-invariant and task-specific information, but it is unimodal and can not be directly transferred to multimodal applications. Particularly, \citet{wang2022s} propose S-Prompts which is multimodal, but this prompting method ignores the modality-level information. Moreover, S-Prompts is a CLIP-based~\citep{radford2021learning} approach which is a language-image scheme and thus can not address problems which has more modalities. In contrast, our proposed method has both modality interaction and task interaction strategies and can be easily transferred to any backbones.

\section{Proposed Method}
\subsection{Problem Formulation}
In real-world dynamic environments, the new data come continually which could have different modality cases. Therefore, we can consider it as a domain-incremental learning task. In a common domain-incremental learning setting, training samples of different domains arrive in sequence (\textit{i.e.} data with different missing modality cases in our task). We denote the sequential datasets as $\mathcal D=\{\mathcal D_1,\mathcal D_2, \cdots, \mathcal D_T \}$, where $\mathcal D_t=\{(\boldsymbol{x_i^t},y_i^t)\}_{i=1}^{N_t}$ represents the dataset for the $t$-th task with $N_t$ training samples. For example, as shown in Table~\ref{denotation}, $\mathcal D_4$ represents the dataset with audio modality missing. In this paper, we consider a case of $M=3$ modalities (audio, video and text) for simplicity. Therefore, $\boldsymbol{x_i^t}$ consists of three modalities and there are a total of $2^M-1=7$ different missing modality cases (shown in Table~\ref{denotation}).

\begin{table}
  \centering
  \caption{The seven different missing modality cases and their denotations.}
  \resizebox{0.93\columnwidth}{!}{
  \begin{tabular}{ccl}\toprule
    \textbf{No.} & \textbf{\{available, missing\}} & \textbf{denotation}\\
    \midrule
    1 & $\{(a, v, t), ()\}$ & $\boldsymbol{x}=(x^a, x^v, x^t)$\\
    2 & $\{(a, v), (t)\}$ & $\boldsymbol{x}=(x^a, x^v, x^{tm})$\\
    3 & $\{(a, t), (v)\}$ & $\boldsymbol{x}=(x^a, x^{vm}, x^t)$\\
    4 & $\{(v, t), (a)\}$ & $\boldsymbol{x}=(x^{am}, x^v, x^t)$\\
    5 & $\{(a), (v, t)\}$ & $\boldsymbol{x}=(x^a, x^{vm}, x^{tm})$\\
    6 & $\{(v), (a, t)\}$ & $\boldsymbol{x}=(x^{am}, x^v, x^{tm})$\\
    7 & $\{(t), (a, v)\}$ & $\boldsymbol{x}=(x^{am} x^{vm}, x^t)$\\
    \bottomrule
  \end{tabular}}
  \label{denotation}
  \vskip -0.1 in
\end{table}

\subsection{Prompt Design}\label{s32}
Existing methods address missing modalities mainly by utilizing complicated modules to generate missing information~\cite{zhao-etal-2021-missing,Du_2018} or aligning multimodal representations~\cite{pham2020translation}. Besides, existing continual methods often cause privacy issues or scalability issues. Motivated by prompt learning methods~\cite{wang2022dualprompt, lee2023multimodal, guo-etal-2024-multimodal}, we propose three types of prompts to address missing modality cases in a continual setting, which is simple and computationally efficient. Specifically, our proposed method contains three types of prompts: modality-specific prompts, task-aware prompts, and task-specific prompts (as shown in Figure~\ref{overall}).

\noindent\textbf{Modality-specific Prompts.} Existing prompt-based methods~\cite{wang2022s, wang2022learning, wang2022dualprompt} mainly focus on task interaction while ignoring interactions between modalities. Therefore, to model inter-modality features, we propose modality-specific prompts. We denote modality-specific prompts as $P_{MS}\in\mathbb R^{M\times\ell\times d}$ where $\ell$ and $d$ represent the length and dimension of the prompt respectively. $P_{MS}$ consists of $P_{MS}^a$, $P_{MS}^v$ and $P_{MS}^t$, which represent audio, video and text modality, respectively. The modality-specific prompts are modality-specific but task-shared. We attach this kind of prompt to the following $n_1$ multi-head self-attention (MSA) layers after the feature projection layer, where $n_1$ is a hyperparameter. The process of attaching prompts to the $i$-th MSA layer is:
\begin{equation}
  \begin{aligned}\label{e1}
  h^{(i)}_a&=\texttt{A-MSA}^{\textrm{(i)}}([P_{MS}^a;h^{(i-1)}_a])\\
  h^{(i)}_v&=\texttt{V-MSA}^{\textrm{(i)}}([P_{MS}^v;h^{(i-1)}_v])\\
  h^{(i)}_t&=\texttt{T-MSA}^{\textrm{(i)}}([P_{MS}^t;h^{(i-1)}_t])
  \end{aligned}
\end{equation}
where $[\cdots]$ is the concatenation operation along the sequence, $h_m^{i}$ is the feature representation of modality $m$ after the $i$-th MSA layer, and $\texttt{A-MSA}^{\textrm{(i)}}$, $\texttt{V-MSA}^{\textrm{(i)}}$ and $\texttt{T-MSA}^{\textrm{(i)}}$ represent the $i$-th audio, video and text MSA layer, respectively.

\noindent\textbf{Task-aware Prompts.} Given the input $x$, the model should be informed of the missing condition of $x$ to address missing information. Therefore, we propose task-aware prompts to learn the inter-modality features between the missing modalities and available modalities. To generate task-aware prompts, we introduce missing keys which are a sign of whether a modality is missing or not. Specifically, we denote missing keys as $K=\{K_m,K_u\}$ where $K_m$ represents a modality is missing while $K_u$ represents a modality is available. $K_m,K_u\in \mathbb R^d$, which are also trainable parameters. Concretely, we use the following equations to generate task-aware prompts for each modality:
\begin{equation}\label{e2}
  P_k=\beta_k\cdot K_m\odot P_{MS}^k + (1-\beta_k)\cdot K_u\odot P_{MS}^k
\end{equation}
where $\odot$ is the element-wise multiplication of the broadcasted vector and the matrix and $P_{MS}^{k}$ are the modality-specific prompts. $k\in\{a,v,t\}$, $\beta_k\in\{0,1\}$ is a sign function to denote whether the modality $k$ is missing. $\beta_k=0$ represents the modality $k$ is missing and $\beta_k=1$ represents the modality $k$ is available. It is worth noting that modality-specific prompts and missing keys are both task-agnostic, but their combinations are task-dependent. This design can not only reduce the number of trainable parameters but also connect the intra- and inter-modality features. Then, we can obtain the task-aware prompts $P_{TA}$ as follows:
\begin{equation}\label{e3}
  P_{TA}=P_a+P_v+P_t
\end{equation}

After we obtain the task-aware prompts $P_{TA}$, we attach them to the next $n_2$ MSA layers following the first $n_1$ layers which are attached with modality-specific prompts. The prompts are attached to the $i$-th MSA layer as follows:
\begin{equation}
  \begin{aligned}\label{e4}
  h^{(i)}_a&=\texttt{A-MSA}^{\textrm{(i)}}([P_{TA};h^{(i-1)}_a])\\
  h^{(i)}_v&=\texttt{V-MSA}^{\textrm{(i)}}([P_{TA};h^{(i-1)}_v])\\
  h^{(i)}_t&=\texttt{T-MSA}^{\textrm{(i)}}([P_{TA};h^{(i-1)}_t])
  \end{aligned}
\end{equation}

Comparing Eq.(\ref{e1}) and Eq.(\ref{e4}), it is easy to discover that the main difference is that different modality has different prompts in Eq.(\ref{e1}) but all modalities have the same prompts in Eq.(\ref{e4}). That is also the main difference between these two types of prompts. Modality-specific prompts are modality-specific but task-agnostic and task-aware prompts are task-dependent but modality-shared. This indicates that modality-specific prompts focus more on learning intra-modality information and task-aware prompts more on inter-modality information.

\noindent\textbf{Task-specific Prompts.} Although task-aware prompts are different in different tasks, they are generated from modality-specific prompts and missing keys which are shared by all tasks. The main role of task-aware prompts is to help the model learn inter-modality information and inform the model of the missing modality condition. Therefore, modality-specific prompts and task-aware prompts are not able to learn task-specific information to address catastrophic forgetting. Based on this observation, we propose task-specific prompts $P_{TS}=\{P_{TS}^{(1)}, P_{TS}^{(2)},\cdots,P_{TS}^{(T)}\}$ to instruct the model for a specific task and address catastrophic forgetting. Specifically, for every task $t$, we have task-specific prompts $P_{TS}^{(t)}\in\mathbb R^{M\times\ell\times d}$. Moreover, $P_{TS}^{(t)}=\{P_{TS_a}^{(t)}$, $P_{TS_v}^{(t)}$, $P_{TS_t}^{(t)}\}$. We attach the prompts the same as before:
\begin{equation}
  \begin{aligned}
  h^{(i)}_a&=\texttt{A-MSA}^{\textrm{(i)}}([P_{TS_a}^{(t)};h^{(i-1)}_a])\\
  h^{(i)}_v&=\texttt{V-MSA}^{\textrm{(i)}}([P_{TS_v}^{(t)};h^{(i-1)}_v])\\
  h^{(i)}_t&=\texttt{T-MSA}^{\textrm{(i)}}([P_{TS_t}^{(t)};h^{(i-1)}_t])
  \end{aligned}
\end{equation}

As before, we attach the prompts to $n_3$ MSA layers after the $n_1+n_2$ MSA layers which are attached with modality-specific prompts and task-aware prompts.

\subsection{Task Interaction Strategy}\label{s33}
Unlike many common domain-incremental learning tasks where there are no evident relationships between the domains, continual multimodal missing modality task has some implicit relationships between different domains. For example, text and audio are always highly relevant because they are both high-semantic information. Different from text and audio, videos contain facial expressions or gestures which are low-semantic information. Therefore, in the representation space, text often has features very similar to that of audio, but much different from that of video.

Based on this observation, we propose to consider audio prompts and text prompts as very similar instances and make them close together in the representation space while making audio and text prompts far from video prompts. Specifically, we adopt a contrastive scheme for task-aware prompts. We consider task-aware prompts of task $\boldsymbol{x}=(x^a,x^v,x^{tm})$ and task $\boldsymbol{x}=(x^{am},x^v,x^t)$, task $\boldsymbol{x}=(x^a,x^{vm},x^{tm})$ and task $\boldsymbol{x}=(x^{am},x^{vm},x^t)$ as positive pairs and others (except $\boldsymbol{x}=(x^a,x^v,x^t)$) as negative pairs. By doing this, task-aware prompts can learn the correlation between different missing modality cases (\textit{i.e.} different tasks), thus strengthening the inter-task relationship.

We consider a modified \textit{NT-Xent} loss~\cite{Chen2020ASF} as our loss function. Let $\textrm{sim}(\boldsymbol{u},\boldsymbol{v})=\boldsymbol{u}^\top\boldsymbol{v}/||\boldsymbol{u}||||\boldsymbol{v}||$ denote the dot product between $\ell_2$ normalized $\boldsymbol{u}$ and $\boldsymbol{v}$ (\textit{i.e.} cosine similarity). The loss function for a positive example $(i,j)$ is:
\begin{equation}\label{e6}
  \ell_{i,j}=-\log\frac{\exp(\textrm{sim}(\boldsymbol{z}_i, \boldsymbol{z}_j)/\tau)}{\sum_{t=1}^T\mathds 1_{[t\neq i, j]}\exp(\textrm{sim}(\boldsymbol{z}_i, \boldsymbol{z}_t)/\tau)}
\end{equation}
where $\tau$ is a temperature parameter and $\mathds 1_{[t\neq i, j]}\in \{0,1\}$ is a sign function evaluating to 1 if $t\neq i, j$. We take the average value of task-aware prompts along the sequence length dimension as $\boldsymbol{z}$. As shown in Table~\ref{denotation}, we denote the index of task-aware prompts of task $\boldsymbol{x}=(x^a,x^v,x^{tm})$, task $\boldsymbol{x}=(x^{am},x^v,x^t)$, task $\boldsymbol{x}=(x^a,x^{vm},x^{tm})$, and task $\boldsymbol{x}=(x^{am},x^{vm},x^t)$ as $2,4,5,7$, respectively. Therefore, we can define our contrastive loss as:
\begin{equation}
  \mathcal L_{con}=\ell_{2,4} + \lambda_2\ell_{5,7}
\end{equation}
where $\lambda_2$ is a trade-off between the two losses.

\subsection{Overall Objective}
We combine the task loss with contrastive loss as:
\begin{equation}
  \mathcal L=\mathcal L_{task}(\hat{y}(\boldsymbol{x}), y) + \lambda_1 \mathcal L_{con}
\end{equation}
where $\hat{y}(\boldsymbol{x})$ is the network prediction, $y$ is the label, $\lambda_1$ is a hyperparameter to balance the two losses, and $\mathcal L_{task}$ is the task-specific loss, \textit{e.g.} cross-entropy loss, L2 loss.

\begin{table*}[h!]
  \centering
  \caption{Quantitative results on CMU-MOSI, IEMOCAP and CH-SIMS datasets. \textbf{Bold}: best exemplar-free results. \underline{Underline}: second best exemplar-free results. * denotes best replay-based results. Lowerbound: training the backbone without any prompts on the continual datasets. Upperbound: supervised finetuning on the i.i.d data of all tasks. Upperbound (ours): supervised finetuning with modality-specific prompts and task-aware prompts on the i.i.d data of all tasks. AA: average accuracy, FM: forgetting measure.}
  \resizebox{\linewidth}{!}{
  \begin{tabular}{l|c|cc|c|cc|c|cc}\toprule
    \multirow{2}{*}{\textbf{Method}} & \multirow{2}{*}{\textbf{Buffer size}} & \multicolumn{2}{c}{\textbf{CMU-MOSI}} & \multirow{2}{*}{\textbf{Buffer size}} & \multicolumn{2}{c}{\textbf{IEMOCAP}}& \multirow{2}{*}{\textbf{Buffer size}} & \multicolumn{2}{c}{\textbf{CH-SIMS}}\\
    \cmidrule(lr){3-4}\cmidrule(lr){6-7}\cmidrule(lr){9-10}
    && AA ($\uparrow$) & FM ($\downarrow$) && AA ($\uparrow$) & FM ($\downarrow$)&& AA ($\uparrow$) & FM ($\downarrow$)\\\midrule
    iCaRL~\citep{rebuffi2017icarl} &\multirow{4}{*}{250}&64.12&3.49&\multirow{4}{*}{500}&54.63&6.11&\multirow{4}{*}{250}&63.79&3.17\\
    A-GEM~\citep{AGEM}&&63.18&4.10&&52.97&7.89&&62.01&4.27\\
    ER~\citep{chaudhry2019tiny}&&65.78&3.44&&57.14&5.09&&65.85&2.76\\
    DER++~\citep{buzzega2020dark}&&64.62&2.74&&54.87&4.50&&63.51&2.96\\
    \midrule
    iCaRL~\citep{rebuffi2017icarl} &\multirow{4}{*}{500}&66.81&2.01&\multirow{4}{*}{1000}&56.46&2.39&\multirow{4}{*}{500}&65.97&1.84\\
    A-GEM~\citep{AGEM}&&65.12&2.88&&54.07&5.78&&65.18&3.01\\
    ER~\citep{chaudhry2019tiny}&&68.91*&1.12&&58.89*&2.98&&68.46*&0.97*\\
    DER++~\citep{buzzega2020dark}&&67.02&0.69*&&57.56&2.40*&&66.18&1.02\\
    \midrule
    EWC~\citep{kirkpatrick2017overcoming}&\multirow{5}{*}{0}&66.44&1.75&\multirow{5}{*}{0}&58.96&2.12&\multirow{5}{*}{0}&65.11&2.04\\
    LwF~\citep{li2017learning}&&64.56&2.97&&54.95&4.87&&63.70&3.09\\
    L2P~\citep{wang2022learning}&&63.79&2.67&&55.68&4.73&&63.61&2.51\\
    DualPrompt~\citep{wang2022dualprompt}&&67.23&\underline{0.73}&&58.15&\underline{1.29}&&\underline{68.73}&\underline{0.89}\\
    S-Prompts~\citep{wang2022s}&&64.83&3.57&&54.30&5.09&&64.96&2.94\\
    \midrule
    MMIM~\cite{han-etal-2021-improving}&\multirow{4}{*}{-}&64.25&5.31&\multirow{4}{*}{-}&52.38&9.15&\multirow{4}{*}{-}&61.37&6.96\\
    MISA~\cite{10.1145/3394171.3413678}&&61.63&6.75&&49.33&10.51&&59.12&7.01\\
    UniMSE~\cite{hu-etal-2022-unimse}&&64.97&5.26&&52.89&9.23&&53.46&6.21\\
    \midrule
    MCTN~\cite{pham2020translation}&\multirow{4}{*}{-}&63.35&4.17&\multirow{4}{*}{-}&56.13&5.35&\multirow{4}{*}{-}&63.11&3.94\\
    MMIN~\cite{zhao-etal-2021-missing}&&65.31&3.92&&56.41&4.36&&64.85&3.11\\
    IPD~\cite{10.1145/3581783.3612291}&&67.13&1.84&&57.63&2.91&&67.16&1.41\\
    MPLMM~\cite{guo-etal-2024-multimodal}&&\underline{70.35}&2.18&&\underline{60.32}&2.90&&68.24&2.77\\
    \midrule
    \textbf{Ours}&\multirow{4}{*}{-}&\textbf{71.87}&\textbf{-0.15}&\multirow{4}{*}{-}&\textbf{62.24}&\textbf{0.08}&\multirow{4}{*}{-}&\textbf{71.11}&\textbf{0.04}\\
    Lowerbound&&62.34&6.18&&51.15&10.31&&61.18&6.87\\
    Upperbound&&71.19&-&&61.74&-&&70.08&-\\
    Upperbound (Ours)&&73.20&-&&64.22&-&&71.98&-\\
    \bottomrule
  \end{tabular}}
  \label{mainresult}
\end{table*}

\section{Experiments}
\subsection{Datasets and Evaluation Metrics}
We validate our methods on CMU-MOSI, IEMOCAP and CH-SIMS.

\noindent\textbf{CMU-MOSI}~\citep{7742221} is a popular dataset for multimodal (audio, text and video) sentiment analysis, comprising 93 English YouTube videos which are carefully selected and divided into 2,199 segments. Each segment is manually annotated with a sentiment score ranging from strongly negative to strongly positive (-3 to +3). 

\noindent\textbf{IEMOCAP}~\citep{Busso2008IEMOCAPIE} contains recorded videos from ten actors in five dyadic conversation sessions. There are different types of emotions (happiness, anger, sadness, frustration and neutral state). In our task, four emotions (happiness, anger, sadness and neutral state) are selected for classfication.

\noindent\textbf{CH-SIMS}~\cite{yu-etal-2020-ch} is a Chinese multimodal sentiment analysis dataset. It contains 2,281 refined video segments in the wild annotated with a sentiment score ranging from strongly negative to strongly positive (-1 to 1). The dataset covers a total number of 474 distinct speakers.

For evaluation, we use Average accuracy (AA) and Forgetting measure (FM). AA is the average accuracy of all tasks and calculated as $AA=\frac{1}{n}\sum_{i=1}^na_{i,n}$ where $a_{i,n}$ is the accuracy on task $i$ after training the model on task $n$. FM measures the performance degradation and is calculated as $FM=\frac{1}{n-1}\sum_{i=1}^{n-1}\max_{j\in[i,n-1]}(a_{i,j}-a_{i,n})$.

\subsection{Baselines}
\noindent\textbf{Continual methods.} They include non-prompting methods: iCaRL \citep{rebuffi2017icarl}, EWC~\citep{kirkpatrick2017overcoming}, LwF~\citep{li2017learning}, A-GEM~\citep{AGEM}, ER~\citep{chaudhry2019tiny}, DER++~\citep{buzzega2020dark}, and prompting methods: L2P~\citep{wang2022learning}, DualPrompt~\citep{wang2022dualprompt}, S-Prompts~\citep{wang2022s}.
For replay-based methods iCaRL, A-GEM, ER, DER++, we use two different replay buffer sizes (250, 500 for CMU-MOSI and CH-SIMS and 500, 1000 for IEMOCAP). 

\noindent\textbf{Robust Multimodal Methods.} Besides, we compare our method with some state-of-the-art multimodal backbones: MISA~\cite{10.1145/3394171.3413678}, MMIM~\cite{han-etal-2021-improving}, UniMSE~\cite{hu-etal-2022-unimse}. We replace the missing modalities with zero vectors.

\noindent\textbf{Missing modality methods.} Moreover, we compare our method with a series of missing modality approaches: MCTN~\cite{pham2020translation}, MMIN~\cite{zhao-etal-2021-missing}, IPD~\cite{10.1145/3581783.3612291}, MPLMM~\cite{guo-etal-2024-multimodal}.

\subsection{Implementation Details}
For fair comparison, we use the multimodal transformer as backbone for continual learning methods. For our proposed method, the dimension $d$ of all the prompts is set to 30 and the length $\ell$ is set to 16 by default. We set $n_1=2$, $n_2=3$ and $n_3=3$. We use L1 loss for CMU-MOSI and CH-SIMS and cross-entropy loss for IEMOCAP. After hyperparameter searching, we set $\lambda_1=0.1$ and $\lambda_2=1$. In all experiments, we use Adam optimizer with a batch size of 64. For other hyperparameters, we follow the original paper of comparing methods. We train all the models for 30 epochs with a learning rate of $1\times 10^{-3}$.

For non-prompting methods iCaRL, EWC, LwF, A-GEM, ER, DER++, we do not freeze the backbone. For prompt-based methods L2P, DualPrompt, S-Prompts and MPLMM, we freeze the pre-trained backbone and only finetune the learnable prompts. 

\subsection{Main Results}
Table~\ref{mainresult} presents the performance of all methods on CMU-MOSI, IEMOCAP and CH-SIMS datasets. 

\noindent\textbf{Comparison with continual learning methods.} Compared with replay-based methods which could lead to privacy issues, our method does not use any buffered data and still can achieve better performance than those with a memory buffer. Compared with exemplar-free continual methods, our method achieves better average results and forgetting measure, indicating the effectiveness of our proposed prompts which promote the model to learn intra-modality, inter-modality and inter-task information. 

\noindent\textbf{Comparison with multimodal and missing modality methods.} Besides, we compare our methods with multimodal and missing modality approaches. The results reveal that multimodal methods all have low average accuracy and high forgetting measure, which indicates that they are not able to deal with missing modality issues and catastrophic forgetting. In comparison, missing modality approaches can achieve comparable or even higher average accuracy than those continual methods due to modules or strategies that are designed to address missing modalities. However, compared to continual methods, these methods often have higher forgetting measure, indicating that they fail to address the catastrophic forgetting. In contrast, our method can not only address the missing modality issue but also deal with catastrophic forgetting in the dynamic environment.

\noindent\textbf{Performance of our method.} We get a negative forgetting measure on CMU-MOSI, which indicates that in the process of learning new tasks, the model performs even better on previous tasks. This demonstrates the effectiveness of our novel design of prompts and task interaction strategy, which enables the model to learn better intra-modality and intra-task relationships, thus making it perform better on previous tasks without forgetting.

Moreover, our proposed method outperforms upperbound slightly which is trained on the i.i.d data of all tasks. This fully demonstrates that our method improves inter-modality communication. Comparing the upperbound and the upperbound using our designed prompts, we could also discover our method makes it easier for the model to learn intra-modality and inter-modality information.

\noindent\textbf{Efficiency of our method.} It is worth noting that the number of trainable parameters of our method only accounts for about 2-3\% of the parameters of the backbone network. With such few parameters, our method can achieve better results than other baseline methods, which indicates that our method is parameter-efficient and effective.

\begin{table}
  \centering
  \caption{An ablation study of three different types of prompts on CMU-MOSI.}
  \resizebox{0.68\columnwidth}{!}{
  \begin{tabular}{ccccc}\toprule
    $P_{MS}$ & $P_{TA}$ & $P_{TS}$ & AA ($\uparrow$) &FM ($\downarrow$)\\
    \midrule
    &&&62.34&6.18\\
    \midrule
    $\checkmark$&&&64.07&4.01\\
    &$\checkmark$&&66.21&3.24\\
    &&$\checkmark$&68.01&1.27\\
    \midrule
    $\checkmark$&$\checkmark$&&69.16&2.08\\
    $\checkmark$&&$\checkmark$&70.34&0.74\\
    &$\checkmark$&$\checkmark$&\underline{70.51}&\underline{0.31}\\
    \midrule
    $\checkmark$&$\checkmark$&$\checkmark$&\textbf{70.91}&\textbf{0.13}\\
    \bottomrule
  \end{tabular}}
  \label{ab1}
\end{table}

\begin{table}
  \centering
  \caption{An ablation study on the benefit of task interaction strategy on CMU-MOSI.}
  \resizebox{0.74\columnwidth}{!}{
  \begin{tabular}{lcc}\toprule
  Method &AA ($\uparrow$)&FM ($\downarrow$)\\
  \midrule
  w/o. task interaction &70.91&0.13\\
  w/o. $\ell_{2,4}$, $\lambda_1=0.1$ &71.14&0.09\\
  w/o. $\ell_{5,7}$, $\lambda_1=0.1$ &71.42&0.06\\
  \midrule
  $\lambda_1=0.2,\lambda_2=1$&71.52&0.04\\
  $\lambda_1=0.1,\lambda_2=2$&71.69&0.01\\
  $\lambda_1=0.1,\lambda_2=0.5$&71.58&-0.04\\
  \midrule
  $\lambda_1=0.1,\lambda_2=1$&\textbf{71.87}&\textbf{-0.15}\\
  \bottomrule
  \end{tabular}}
  \label{ab2}
\end{table}

\subsection{Ablation Study}
\noindent\textbf{Effectiveness of three types of prompts.} In Table~\ref{ab1}, we show quantitative results of the benefits of three types of prompts. It is easy to find that task-specific prompts contribute most to addressing catastrophic forgetting. As shown in the table, the forgetting measure of the model with only task-specific prompts $P_{TS}$ is lower than that of the model with modality-specific prompts $P_{MS}$ and task-aware prompts $P_{TA}$. This indicates that task-specific prompts help the model learn intra-task information, which plays a very important role in dealing with forgetting. Besides, modality-specific prompts and task-aware prompts help a lot in improving the model's average accuracy. Modality-specific prompts help the model learn intra-modality information and task-aware prompts help the model learn inter-modality and inter-task information. The combination of three types of prompts further enhances the performance of the model, which fully convinces us of the effectiveness of our proposed prompts.

\begin{figure}
  \centering
  \includegraphics[width=0.7\linewidth]{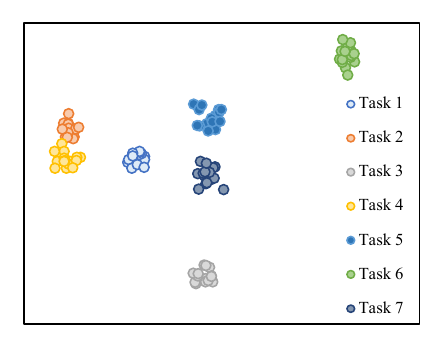}
  \caption{t-SNE visualization of task-aware prompts on the CMU-MOSI dataset. Each point represents a prompt vector. Tasks 1-7 are shown in Table~\ref{denotation}.}
  \label{tsne}
\end{figure}

Moreover, we visualize task-aware prompts using t-SNE in Figure~\ref{tsne}. We can observe that points of Task 2 and Task 3, Task 5 and Task 7 are very close to each other. This indicates the effectiveness of our task interaction strategy, which helps the model learn inter-task relationships. Besides, task-aware prompts of different tasks are well-separated, which demonstrates that these prompts help the model learn task-dependent knowledge.

\noindent\textbf{Effectiveness of task interaction strategy.} In Section~\ref{s33}, we introduce a task interaction strategy. To demonstrate the effectiveness of our proposed task interaction strategy, we present our ablation results in Table~\ref{ab2}. We find that the model with two loss terms performs much better on average accuracy and forgetting measure than that without the loss terms. Besides, the model without task interaction strategy performs worse than the upperbound method shown in Table~\ref{mainresult}. This indicates that our proposed task interaction strategy acts as a bridge between tasks and helps the model learn the inter-task information, thus outperforming the upperbound method. In the fourth to sixth rows of the table, we explore the impact of the trade-off terms $\lambda_1$ and $\lambda_2$ on the performance of the model. The results reveal that the performance of the model is not sensitive to the value of $\lambda_1$ and $\lambda_2$.

\begin{table}
  \centering
  \caption{An ablation study of the sequence of attaching these three types of prompts on CMU-MOSI. $A\rightarrow B\rightarrow C$ represents that we attach A prompts at the first $n_1$ MSA layers, B prompts at the following $n_2$ MSA layers, and C prompts at the next $n_3$ MSA layers. Here, we set $n_1=2$, $n_2=3$, $n_3=3$. \textbf{Bold}: best results.}
  \resizebox{0.82\linewidth}{!}{
  \begin{tabular}{ccc}\toprule
    \textbf{Prompt Sequence}&AA ($\uparrow$)&FM ($\downarrow$)\\
    \midrule
    $P_{MS}\rightarrow P_{TA}\rightarrow P_{TS}$&\textbf{71.87}&\textbf{-0.15}\\
    $P_{MS}\rightarrow P_{TS}\rightarrow P_{TA}$&70.96&-0.03\\
    $P_{TA}\rightarrow P_{MS}\rightarrow P_{TS}$&71.23&0.06\\
    $P_{TA}\rightarrow P_{TS}\rightarrow P_{MS}$&70.57&0.04\\
    $P_{TS}\rightarrow P_{MS}\rightarrow P_{TA}$&70.00&0.10\\
    $P_{TS}\rightarrow P_{TA}\rightarrow P_{MS}$&69.98&0.28\\
    \bottomrule
  \end{tabular}}
  \label{ab3}
\end{table}

\begin{table}
  \centering
  \caption{An ablation study of the specific positions of prompts on the CMU-MOSI dataset. Our backbone has ten MSA layers in total. \textbf{Bold}: best results.}
  \resizebox{0.9\linewidth}{!}{
  \begin{tabular}{lllcc}\toprule
    $P_{MS}$ &$P_{TA}$&$P_{TS}$&AA ($\uparrow$)&FM ($\downarrow$)\\
    \midrule
    $[1, 2]$&$[3, 4, 5]$&$[6, 7, 8, 9]$&70.06&0.23\\
    $[1, 2]$&$[3, 4, 5]$&$[6, 7, 8]$&\textbf{71.87}&\textbf{-0.15}\\
    $[1, 2]$&$[3, 4, 5]$&$[6, 7]$&71.10&0.07\\
    $[1, 2, 3]$&$[4, 5]$&$[6, 7, 8]$&71.42&-0.09\\
    \bottomrule
  \end{tabular}
  }
  \label{ab4}
\end{table}

\noindent\textbf{Exploration of where to attach prompts.} We first conduct a series of experiments to explore the sequence of three types of prompts and present our results in Table~\ref{ab3}. We can find that the model with task-aware prompts $P_{TA}$ in front of task-specific prompts $P_{TS}$ always outperforms the model with $P_{TA}$ behind $P_{TS}$. This indicates that compared to task-specific prompts, task-aware prompts learn low-level features, serving as a guideline to task-specific prompts and helping task-specific prompts learn better intra-task information. Besides, modality-specific prompts instruct the model to learn intra-modality information which are low-level features at early stages. Therefore, modality-specific prompts should be placed in front of the other two types of prompts.

Furthermore, we explore the specific positions of these prompts and present our results in Table~\ref{ab4}. Comparing the results in the first row and the second row in the table, we find that it is not opportune to attach prompts at the back layers of the network. The highest performance demonstrates the effectiveness of our design of prompts.

\begin{figure}
  \centering
  \includegraphics[width=\linewidth]{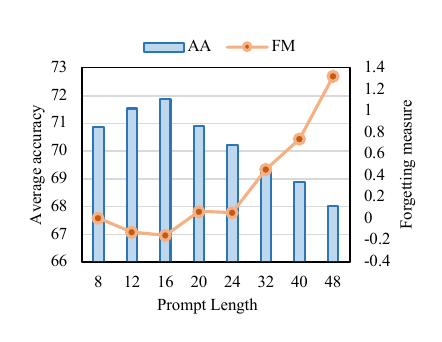}
  \caption{Quantitative results on the CMU-MOSI dataset with different prompt lengths $\ell$.}
  \label{pl}
\end{figure}

\noindent\textbf{Impact of the length of prompts.} To study the impact of prompt length on our model, we train our model on CMU-MOSI with eight different prompt lengths and present results in Figure~\ref{pl}. Intuitively, the longer the prompt length, the better the performance of the model. However, as the results are shown in the figure, we find that with the increasing length $\ell$, the performance first improves and then declines with the peak performance at $\ell=16$. This suggests that our proposed method can achieve great results with a relatively small number of parameters.

\section{Conclusion}
In this paper, we introduce the task of \textit{continual multimodal missing modality} to tackle the challenges posed by missing modalities in dynamic environments. We propose a novel and efficient prompt design consisting of three distinct types of prompts, complemented by a contrastive task interaction strategy aimed at mitigating catastrophic forgetting in the multimodal domain. Our approach facilitates effective learning of intra-modality, inter-modality, intra-task, and inter-task features, enhancing the model's adaptability. Extensive experiments and ablation studies validate the robustness and efficacy of our proposed method. Given that cases of missing modalities frequently arise during data collection in real-world scenarios, we believe our approach represents a significant step towards practical applications in multimodal fields facing ongoing missing modality challenges. 

\section*{Limitations}
In our approach, the number of task-specific prompts is the same as the number of tasks. However, the number of tasks increases exponentially as the number of modalities increases. Therefore, when there are many modalities, it would cost large computational resources. Beyond this work, we believe some promising future works would solve this problem.

\bibliography{anthology,custom}

\begin{thebibliography}{44}
\expandafter\ifx\csname natexlab\endcsname\relax\def\natexlab#1{#1}\fi

\bibitem[{Aljundi et~al.(2018)Aljundi, Babiloni, Elhoseiny, Rohrbach, and Tuytelaars}]{aljundi2018memory}
Rahaf Aljundi, Francesca Babiloni, Mohamed Elhoseiny, Marcus Rohrbach, and Tinne Tuytelaars. 2018.
\newblock Memory aware synapses: Learning what (not) to forget.
\newblock In \emph{Proceedings of the European conference on computer vision (ECCV)}, pages 139--154.

\bibitem[{Busso et~al.(2008)Busso, Bulut, Lee, Kazemzadeh, Provost, Kim, Chang, Lee, and Narayanan}]{Busso2008IEMOCAPIE}
Carlos Busso, Murtaza Bulut, Chi-Chun Lee, Ebrahim~(Abe) Kazemzadeh, Emily~Mower Provost, Samuel Kim, Jeannette~N. Chang, Sungbok Lee, and Shrikanth~S. Narayanan. 2008.
\newblock \href {https://api.semanticscholar.org/CorpusID:11820063} {Iemocap: interactive emotional dyadic motion capture database}.
\newblock \emph{Language Resources and Evaluation}, 42:335--359.

\bibitem[{Buzzega et~al.(2020)Buzzega, Boschini, Porrello, Abati, and Calderara}]{buzzega2020dark}
Pietro Buzzega, Matteo Boschini, Angelo Porrello, Davide Abati, and Simone Calderara. 2020.
\newblock Dark experience for general continual learning: a strong, simple baseline.
\newblock \emph{Advances in neural information processing systems}, 33:15920--15930.

\bibitem[{Cai et~al.(2018)Cai, Wang, Gao, Shen, and Ji}]{10.1145/3219819.3219963}
Lei Cai, Zhengyang Wang, Hongyang Gao, Dinggang Shen, and Shuiwang Ji. 2018.
\newblock \href {https://doi.org/10.1145/3219819.3219963} {Deep adversarial learning for multi-modality missing data completion}.
\newblock In \emph{Proceedings of the 24th ACM SIGKDD International Conference on Knowledge Discovery \& Data Mining}, KDD '18, page 1158–1166, New York, NY, USA. Association for Computing Machinery.

\bibitem[{Cha et~al.(2021)Cha, Lee, and Shin}]{cha2021co2l}
Hyuntak Cha, Jaeho Lee, and Jinwoo Shin. 2021.
\newblock Co2l: Contrastive continual learning.
\newblock In \emph{Proceedings of the IEEE/CVF International conference on computer vision}, pages 9516--9525.

\bibitem[{Chaudhry et~al.(2019{\natexlab{a}})Chaudhry, Ranzato, Rohrbach, and Elhoseiny}]{AGEM}
Arslan Chaudhry, Marc’Aurelio Ranzato, Marcus Rohrbach, and Mohamed Elhoseiny. 2019{\natexlab{a}}.
\newblock Efficient lifelong learning with a-gem.
\newblock In \emph{ICLR}.

\bibitem[{Chaudhry et~al.(2019{\natexlab{b}})Chaudhry, Rohrbach, Elhoseiny, Ajanthan, Dokania, Torr, and Ranzato}]{chaudhry2019tiny}
Arslan Chaudhry, Marcus Rohrbach, Mohamed Elhoseiny, Thalaiyasingam Ajanthan, Puneet~K. Dokania, Philip H.~S. Torr, and Marc'Aurelio Ranzato. 2019{\natexlab{b}}.
\newblock \href {http://arxiv.org/abs/1902.10486} {On tiny episodic memories in continual learning}.

\bibitem[{Chen et~al.(2020)Chen, Kornblith, Norouzi, and Hinton}]{Chen2020ASF}
Ting Chen, Simon Kornblith, Mohammad Norouzi, and Geoffrey~E. Hinton. 2020.
\newblock A simple framework for contrastive learning of visual representations.
\newblock In \emph{International conference on machine learning}.

\bibitem[{Du et~al.(2018)Du, Du, Wang, Li, Zheng, Lu, and He}]{Du_2018}
Changde Du, Changying Du, Hao Wang, Jinpeng Li, Wei-Long Zheng, Bao-Liang Lu, and Huiguang He. 2018.
\newblock \href {https://doi.org/10.1145/3240508.3240528} {Semi-supervised deep generative modelling of incomplete multi-modality emotional data}.
\newblock In \emph{Proceedings of the 26th {ACM} international conference on Multimedia}. {ACM}.

\bibitem[{Ebrahimi et~al.(2020)Ebrahimi, Meier, Calandra, Darrell, and Rohrbach}]{ebrahimi2020adversarial}
Sayna Ebrahimi, Franziska Meier, Roberto Calandra, Trevor Darrell, and Marcus Rohrbach. 2020.
\newblock Adversarial continual learning.
\newblock In \emph{Computer Vision--ECCV 2020: 16th European Conference, Glasgow, UK, August 23--28, 2020, Proceedings, Part XI 16}, pages 386--402. Springer.

\bibitem[{Guo et~al.(2024{\natexlab{a}})Guo, Cheng, Wu, and Jin}]{guo2024wander}
Zirun Guo, Xize Cheng, Yangyang Wu, and Tao Jin. 2024{\natexlab{a}}.
\newblock A wander through the multimodal landscape: Efficient transfer learning via low-rank sequence multimodal adapter.
\newblock \emph{arXiv preprint arXiv:2412.08979}.

\bibitem[{Guo and Jin(2025)}]{guo2025smoothing}
Zirun Guo and Tao Jin. 2025.
\newblock \href {https://openreview.net/forum?id=rObkvzJxTG} {Smoothing the shift: Towards stable test-time adaptation under complex multimodal noises}.
\newblock In \emph{The Thirteenth International Conference on Learning Representations}.

\bibitem[{Guo et~al.(2024{\natexlab{b}})Guo, Jin, Chen, and Zhao}]{guo2024classifier}
Zirun Guo, Tao Jin, Jingyuan Chen, and Zhou Zhao. 2024{\natexlab{b}}.
\newblock Classifier-guided gradient modulation for enhanced multimodal learning.
\newblock In \emph{The Thirty-eighth Annual Conference on Neural Information Processing Systems}.

\bibitem[{Guo et~al.(2024{\natexlab{c}})Guo, Jin, and Zhao}]{guo-etal-2024-multimodal}
Zirun Guo, Tao Jin, and Zhou Zhao. 2024{\natexlab{c}}.
\newblock \href {https://doi.org/10.18653/v1/2024.acl-long.94} {Multimodal prompt learning with missing modalities for sentiment analysis and emotion recognition}.
\newblock In \emph{Proceedings of the 62nd Annual Meeting of the Association for Computational Linguistics (Volume 1: Long Papers)}, pages 1726--1736, Bangkok, Thailand. Association for Computational Linguistics.

\bibitem[{Han et~al.(2021)Han, Chen, and Poria}]{han-etal-2021-improving}
Wei Han, Hui Chen, and Soujanya Poria. 2021.
\newblock \href {https://doi.org/10.18653/v1/2021.emnlp-main.723} {Improving multimodal fusion with hierarchical mutual information maximization for multimodal sentiment analysis}.
\newblock In \emph{Proceedings of the 2021 Conference on Empirical Methods in Natural Language Processing}, pages 9180--9192, Online and Punta Cana, Dominican Republic. Association for Computational Linguistics.

\bibitem[{Hazarika et~al.(2020)Hazarika, Zimmermann, and Poria}]{10.1145/3394171.3413678}
Devamanyu Hazarika, Roger Zimmermann, and Soujanya Poria. 2020.
\newblock \href {https://doi.org/10.1145/3394171.3413678} {Misa: Modality-invariant and -specific representations for multimodal sentiment analysis}.
\newblock In \emph{Proceedings of the 28th ACM International Conference on Multimedia}, MM '20, page 1122–1131, New York, NY, USA. Association for Computing Machinery.

\bibitem[{Hu et~al.(2021)Hu, Shen, Wallis, Allen-Zhu, Li, Wang, Wang, and Chen}]{hu2021lora}
Edward~J Hu, Yelong Shen, Phillip Wallis, Zeyuan Allen-Zhu, Yuanzhi Li, Shean Wang, Lu~Wang, and Weizhu Chen. 2021.
\newblock Lora: Low-rank adaptation of large language models.
\newblock \emph{arXiv preprint arXiv:2106.09685}.

\bibitem[{Hu et~al.(2022)Hu, Lin, Zhao, Lu, Wu, and Li}]{hu-etal-2022-unimse}
Guimin Hu, Ting-En Lin, Yi~Zhao, Guangming Lu, Yuchuan Wu, and Yongbin Li. 2022.
\newblock \href {https://doi.org/10.18653/v1/2022.emnlp-main.534} {{U}ni{MSE}: Towards unified multimodal sentiment analysis and emotion recognition}.
\newblock In \emph{Proceedings of the 2022 Conference on Empirical Methods in Natural Language Processing}, pages 7837--7851, Abu Dhabi, United Arab Emirates. Association for Computational Linguistics.

\bibitem[{Jin et~al.(2023)Jin, Cheng, Li, Lin, Wang, and Zhao}]{10.1145/3581783.3612291}
Tao Jin, Xize Cheng, Linjun Li, Wang Lin, Ye~Wang, and Zhou Zhao. 2023.
\newblock \href {https://doi.org/10.1145/3581783.3612291} {Rethinking missing modality learning from a decoding perspective}.
\newblock In \emph{Proceedings of the 31st ACM International Conference on Multimedia}, MM '23, page 4431–4439, New York, NY, USA. Association for Computing Machinery.

\bibitem[{Kirkpatrick et~al.(2017)Kirkpatrick, Pascanu, Rabinowitz, Veness, Desjardins, Rusu, Milan, Quan, Ramalho, Grabska-Barwinska et~al.}]{kirkpatrick2017overcoming}
James Kirkpatrick, Razvan Pascanu, Neil Rabinowitz, Joel Veness, Guillaume Desjardins, Andrei~A Rusu, Kieran Milan, John Quan, Tiago Ramalho, Agnieszka Grabska-Barwinska, et~al. 2017.
\newblock Overcoming catastrophic forgetting in neural networks.
\newblock \emph{Proceedings of the national academy of sciences}, 114(13):3521--3526.

\bibitem[{Lee et~al.(2023)Lee, Tsai, Chiu, and Lee}]{lee2023multimodal}
Yi-Lun Lee, Yi-Hsuan Tsai, Wei-Chen Chiu, and Chen-Yu Lee. 2023.
\newblock Multimodal prompting with missing modalities for visual recognition.
\newblock In \emph{Proceedings of the IEEE/CVF Conference on Computer Vision and Pattern Recognition}, pages 14943--14952.

\bibitem[{Li et~al.(2023)Li, Li, Savarese, and Hoi}]{Li2023BLIP2BL}
Junnan Li, Dongxu Li, Silvio Savarese, and Steven C.~H. Hoi. 2023.
\newblock Blip-2: Bootstrapping language-image pre-training with frozen image encoders and large language models.
\newblock In \emph{International Conference on Machine Learning}.

\bibitem[{Li et~al.(2019)Li, Zhou, Wu, Socher, and Xiong}]{li2019learn}
Xilai Li, Yingbo Zhou, Tianfu Wu, Richard Socher, and Caiming Xiong. 2019.
\newblock Learn to grow: A continual structure learning framework for overcoming catastrophic forgetting.
\newblock In \emph{International Conference on Machine Learning}, pages 3925--3934. PMLR.

\bibitem[{Li and Hoiem(2017)}]{li2017learning}
Zhizhong Li and Derek Hoiem. 2017.
\newblock Learning without forgetting.
\newblock \emph{IEEE transactions on pattern analysis and machine intelligence}, 40(12):2935--2947.

\bibitem[{Lin et~al.(2024)Lin, Chen, Shi, Guo, Zhu, Wang, Jin, Zhao, Wu, YAN, and Zhang}]{lin2024action}
Wang Lin, Jingyuan Chen, Jiaxin Shi, Zirun Guo, Yichen Zhu, Zehan Wang, Tao Jin, Zhou Zhao, Fei Wu, Shuicheng YAN, and Hanwang Zhang. 2024.
\newblock \href {https://openreview.net/forum?id=h2e4G2YiwR} {Action imitation in common action space for customized action image synthesis}.
\newblock In \emph{The Thirty-eighth Annual Conference on Neural Information Processing Systems}.

\bibitem[{Ma et~al.(2021)Ma, Ren, Zhao, Tulyakov, Wu, and Peng}]{ma2021smil}
Mengmeng Ma, Jian Ren, Long Zhao, Sergey Tulyakov, Cathy Wu, and Xi~Peng. 2021.
\newblock Smil: Multimodal learning with severely missing modality.
\newblock In \emph{Proceedings of the AAAI Conference on Artificial Intelligence}, volume~35, pages 2302--2310.

\bibitem[{Mallya and Lazebnik(2018)}]{mallya2018packnet}
Arun Mallya and Svetlana Lazebnik. 2018.
\newblock Packnet: Adding multiple tasks to a single network by iterative pruning.
\newblock In \emph{Proceedings of the IEEE conference on Computer Vision and Pattern Recognition}, pages 7765--7773.

\bibitem[{McCloskey and Cohen(1989)}]{mccloskey1989catastrophic}
Michael McCloskey and Neal~J Cohen. 1989.
\newblock Catastrophic interference in connectionist networks: The sequential learning problem.
\newblock In \emph{Psychology of learning and motivation}, volume~24, pages 109--165. Elsevier.

\bibitem[{Pham et~al.(2019)Pham, Liang, Manzini, Morency, and P{\'o}czos}]{pham2020translation}
Hai Pham, Paul~Pu Liang, Thomas Manzini, Louis-Philippe Morency, and Barnab{\'a}s P{\'o}czos. 2019.
\newblock Found in translation: Learning robust joint representations by cyclic translations between modalities.
\newblock In \emph{Proceedings of the AAAI Conference on Artificial Intelligence}, volume~33, pages 6892--6899.

\bibitem[{Qian et~al.(2023)Qian, Wang, Duan, Qin, Li, and Zhu}]{qian2023decouple}
Zi~Qian, Xin Wang, Xuguang Duan, Pengda Qin, Yuhong Li, and Wenwu Zhu. 2023.
\newblock Decouple before interact: Multi-modal prompt learning for continual visual question answering.
\newblock In \emph{Proceedings of the IEEE/CVF International Conference on Computer Vision}, pages 2953--2962.

\bibitem[{Radford et~al.(2021)Radford, Kim, Hallacy, Ramesh, Goh, Agarwal, Sastry, Askell, Mishkin, Clark et~al.}]{radford2021learning}
Alec Radford, Jong~Wook Kim, Chris Hallacy, Aditya Ramesh, Gabriel Goh, Sandhini Agarwal, Girish Sastry, Amanda Askell, Pamela Mishkin, Jack Clark, et~al. 2021.
\newblock Learning transferable visual models from natural language supervision.
\newblock In \emph{International conference on machine learning}, pages 8748--8763. PMLR.

\bibitem[{Rebuffi et~al.(2017)Rebuffi, Kolesnikov, Sperl, and Lampert}]{rebuffi2017icarl}
Sylvestre-Alvise Rebuffi, Alexander Kolesnikov, Georg Sperl, and Christoph~H Lampert. 2017.
\newblock icarl: Incremental classifier and representation learning.
\newblock In \emph{Proceedings of the IEEE conference on Computer Vision and Pattern Recognition}, pages 2001--2010.

\bibitem[{Rolnick et~al.(2019)Rolnick, Ahuja, Schwarz, Lillicrap, and Wayne}]{rolnick2019experience}
David Rolnick, Arun Ahuja, Jonathan Schwarz, Timothy Lillicrap, and Gregory Wayne. 2019.
\newblock Experience replay for continual learning.
\newblock \emph{Advances in Neural Information Processing Systems}, 32.

\bibitem[{Serra et~al.(2018)Serra, Suris, Miron, and Karatzoglou}]{serra2018overcoming}
Joan Serra, Didac Suris, Marius Miron, and Alexandros Karatzoglou. 2018.
\newblock Overcoming catastrophic forgetting with hard attention to the task.
\newblock In \emph{International conference on machine learning}, pages 4548--4557. PMLR.

\bibitem[{Shin et~al.(2017)Shin, Lee, Kim, and Kim}]{shin2017continual}
Hanul Shin, Jung~Kwon Lee, Jaehong Kim, and Jiwon Kim. 2017.
\newblock Continual learning with deep generative replay.
\newblock \emph{Advances in neural information processing systems}, 30.

\bibitem[{Wang et~al.(2022{\natexlab{a}})Wang, Huang, and Hong}]{wang2022s}
Yabin Wang, Zhiwu Huang, and Xiaopeng Hong. 2022{\natexlab{a}}.
\newblock S-prompts learning with pre-trained transformers: An occam’s razor for domain incremental learning.
\newblock \emph{Advances in Neural Information Processing Systems}, 35:5682--5695.

\bibitem[{Wang et~al.(2022{\natexlab{b}})Wang, Zhang, Ebrahimi, Sun, Zhang, Lee, Ren, Su, Perot, Dy et~al.}]{wang2022dualprompt}
Zifeng Wang, Zizhao Zhang, Sayna Ebrahimi, Ruoxi Sun, Han Zhang, Chen-Yu Lee, Xiaoqi Ren, Guolong Su, Vincent Perot, Jennifer Dy, et~al. 2022{\natexlab{b}}.
\newblock Dualprompt: Complementary prompting for rehearsal-free continual learning.
\newblock \emph{European Conference on Computer Vision}.

\bibitem[{Wang et~al.(2022{\natexlab{c}})Wang, Zhang, Lee, Zhang, Sun, Ren, Su, Perot, Dy, and Pfister}]{wang2022learning}
Zifeng Wang, Zizhao Zhang, Chen-Yu Lee, Han Zhang, Ruoxi Sun, Xiaoqi Ren, Guolong Su, Vincent Perot, Jennifer Dy, and Tomas Pfister. 2022{\natexlab{c}}.
\newblock Learning to prompt for continual learning.
\newblock In \emph{Proceedings of the IEEE/CVF Conference on Computer Vision and Pattern Recognition}, pages 139--149.

\bibitem[{Yan et~al.(2025)Yan, Lin, Guo, Wang, Feng, Yang, Wang, and Jin}]{yan2025diffprompt}
Weicai Yan, Wang Lin, Zirun Guo, Ye~Wang, Fangming Feng, Xiaoda Yang, Zehan Wang, and Tao Jin. 2025.
\newblock Diff-prompt: Diffusion-driven prompt generator with mask supervision.
\newblock In \emph{The Thirteenth International Conference on Learning Representations}.

\bibitem[{Yan et~al.(2024)Yan, Wang, Lin, Guo, Zhao, and Jin}]{yan2024lowrank}
Weicai Yan, Ye~Wang, Wang Lin, Zirun Guo, Zhou Zhao, and Tao Jin. 2024.
\newblock Low-rank prompt interaction for continual vision-language retrieval.
\newblock In \emph{Proceedings of the 32nd ACM International Conference on Multimedia}, pages 8257--8266.

\bibitem[{Yu et~al.(2020)Yu, Xu, Meng, Zhu, Ma, Wu, Zou, and Yang}]{yu-etal-2020-ch}
Wenmeng Yu, Hua Xu, Fanyang Meng, Yilin Zhu, Yixiao Ma, Jiele Wu, Jiyun Zou, and Kaicheng Yang. 2020.
\newblock \href {https://doi.org/10.18653/v1/2020.acl-main.343} {{CH}-{SIMS}: A {C}hinese multimodal sentiment analysis dataset with fine-grained annotation of modality}.
\newblock In \emph{Proceedings of the 58th Annual Meeting of the Association for Computational Linguistics}, pages 3718--3727, Online. Association for Computational Linguistics.

\bibitem[{Zadeh et~al.(2016)Zadeh, Zellers, Pincus, and Morency}]{7742221}
Amir Zadeh, Rowan Zellers, Eli Pincus, and Louis-Philippe Morency. 2016.
\newblock \href {https://doi.org/10.1109/MIS.2016.94} {Multimodal sentiment intensity analysis in videos: Facial gestures and verbal messages}.
\newblock \emph{IEEE Intelligent Systems}, 31(6):82--88.

\bibitem[{Zenke et~al.(2017)Zenke, Poole, and Ganguli}]{zenke2017continual}
Friedemann Zenke, Ben Poole, and Surya Ganguli. 2017.
\newblock Continual learning through synaptic intelligence.
\newblock In \emph{International conference on machine learning}, pages 3987--3995. PMLR.

\bibitem[{Zhao et~al.(2021)Zhao, Li, and Jin}]{zhao-etal-2021-missing}
Jinming Zhao, Ruichen Li, and Qin Jin. 2021.
\newblock \href {https://doi.org/10.18653/v1/2021.acl-long.203} {Missing modality imagination network for emotion recognition with uncertain missing modalities}.
\newblock In \emph{Proceedings of the 59th Annual Meeting of the Association for Computational Linguistics and the 11th International Joint Conference on Natural Language Processing (Volume 1: Long Papers)}, pages 2608--2618, Online. Association for Computational Linguistics.

\end{thebibliography}

\end{document}